\documentclass[oneside,11pt]{article}

\usepackage[total = 6in]{geometry}
\usepackage{tikz}
\usepackage{amsmath}
\usepackage{amsfonts}
\usepackage{amssymb}
\usepackage{algorithm}
\usepackage{algorithmic}
\usepackage{natbib}

\tikzset{
	node/.style={circle,inner sep=1mm,minimum size=0.8cm,draw,
      very thick,black,fill=white,text=black},
	nondirectional/.style={very thick,black},
	unidirectional/.style={nondirectional,shorten >=2pt,-stealth},
	bidirectional/.style={unidirectional,bend right=10}
}

\usepackage[utf8]{inputenc}
\inputencoding{utf8}

\usepackage{hyperref,color,soul,booktabs}
\setulcolor{blue}
\newcommand{\BibTeX}{\textsc{B\kern-0.1emi\kern-0.017emb}\kern-0.15em\TeX}

\newtheorem{remark}{Remark}
\newtheorem{example}{Example}

\begin{document}

\title{Learning DAGs without imposing acyclicity}

\author{Gherardo Varando \footnote{gherardo.varando@math.ku.dk}}
\date{}
\maketitle

\begin{abstract}
We explore if it is possible to learn a directed acyclic 
graph (DAG) from data
without imposing explicitly 
the acyclicity constraint. In particular, for Gaussian
distributions, we frame structural learning as a
sparse matrix factorization problem and we empirically show 
that solving an $\ell_1$-penalized optimization yields 
to good recovery of the true graph and, in general, to 
almost-DAG graphs. Moreover, this approach is computationally 
efficient and is not affected by the explosion of combinatorial   
complexity as in classical structural learning algorithms. 
 \end{abstract}

\section{Introduction}

A vast literature exists on learning DAGs from data. 
Usually algorithms are classified as constrained based or score
based~\citep{scutari2019}. 
Score-based methods optimize some score function over the space of DAGs, usually
employing some heuristic such as greedy search. 
Constraint-based methods, such as the PC algorithm~\citep{spirtes1993, colombo14}, use 
instead conditional independence testing to prune
edges from the graph and apply sets of rules to direct some of the remaining edges 
and find an estimate of the Markov equivalence class~\citep{chickering95}. 
Recently,~\cite{zheng2018, zheng2020} proposed the use of optimization techniques 
 to estimate  DAG 
structures by writing the acyclicity condition of a directed graph as a smooth 
constraints for the weighted adjacency matrix. 
All the methods available in the literature for structure recovery of DAGs 
impose somehow the acyclicity condition. We propose instead to consider the 
estimation of DAGs as a general sparse matrix-factorization problem for the 
inverse covariance matrix arising from linear structural equation
models~\citep{drton2018, spirtes1995, richardson1997}. We estimate such sparse matrix-factorization 
solving an $\ell_1$-penalized minus log-likelihood minimization using a 
straightforward proximal gradient method. 
The proposed method takes inspiration from optimization-based algorithms for 
structure recovery such as the graphical lasso~\citep{friedman2007} and 
especially the method proposed in \cite{varando2020} where 
covariance matrices are parametrized as solutions of, eventually sparse, 
Lyapunov equations. 
In Section~\ref{sec:simulations} we perform a simulation study and observe that 
the proposed method is competitive with classical approaches from the
literature on the recovery of the true graph, while being generally faster. 

A fortran implementation of the method, together with examples of its
usage within R and python, 
 is available at
\href{https://github.com/gherardovarando/nodag}{\texttt{github.com/gherardovarando/nodag}}.

\section{Linear structural equation models}

A linear structural equation 
model (SEM) with independent Gaussian noise is a statistical model for a 
$p$-dimensional random vector $X$ defined as the solution of 
\begin{equation} 
 X = \Lambda^t X + \epsilon, 
	\label{eq:system}
\end{equation}
where we assume that $\epsilon$ is a $p$-dimensional zero-mean independent
Gaussian noise and $\Lambda \in \mathbb{R}^{p\times p}$. 
If $I - \Lambda$ is invertible, 
equation~\eqref{eq:system} implies the covariance parametrization~\citep{drton2018}  
\begin{equation}
	\operatorname{Var}(X) =  (I-\Lambda)^{-t} \Omega  (I- \Lambda)^{-1}    
	\label{eq:sigma}
\end{equation}
where $\Omega = \operatorname{Var}(\epsilon)$ is a diagonal positive 
definite matrix.  
The connection between Gaussian Bayesian networks 
and the system of equations~\eqref{eq:system} is 
immediate if we assume that the matrix $\Lambda$ has a sparsity structure 
compatible with a given directed acyclic graph ${G}$.

Inverting equation~\ref{eq:sigma} we obtain the factorization of the 
inverse covariance matrix as
\begin{equation}
	\operatorname{Var}(X)^{-1} = (I - \Lambda) \Omega^{-1} (I - \Lambda)^t
	= A A ^t, 
	\label{eq:precision} 
\end{equation}
with $A = (I - \Lambda){\Omega}^{-{1}/{2}}$ having the same off-diagonal
sparsity pattern as  $\Lambda$.

Parametrizing the multivariate Gaussian distribution with the inverse covariance
matrix, we can define the linear structural equation model with independent
Gaussian noise, and associate graph ${G}$, 
as the family of normal distributions $\mathcal{N}(\mu, \Sigma)$ with $\Sigma^{-1}$ in
\[ \mathcal{M}_{{G}} = \left\{ AA^t \text{ s.t. } A \in \mathbb{R}^{p\times p} 
\text{ invertible and } A_{ij} = 0 \text{ if } i
\not\rightarrow j \text{ in } {G} \right\} \]
In particular when the graph ${G}$ is acyclic the set of inverse covariance matrices 
$\mathcal{M}_{G}$ corresponds to Gaussian Bayesian network models.   

\subsection{Markov equivalence classes}
\label{sec:markov}

It is known that in general is not possible to recover completely the graph $G$
from observational data, since different graphs give rise to the same
statistical model, in which case we say that the two graphs are
equivalent~\citep{heckerman1994}. 
The equivalence classes obtained considering the quotient of the space of 
directed graphs with respect to the above equivalence are called Markov
equivalence classes. In particular, if the graph $G$ is a DAG 
its Markov equivalence class consists of all DAGs having the same skeleton and
exactly the same v-structures~\citep{andersson1997, heckerman1994}. 
A completely partially directed acyclic graph (CPDAG) 
can be used to represent the Markov
equivalence class for DAGs~\citep{andersson1997}. A CPDAG is a partially  
directed graph where directed edges represent edges that have the same
directions in all DAGs belonging to the Markov class, while undirected edges
are drawn where there exists DAGs in the Markov equivalence class with different 
directions for a given edge. 

\section{Structure recovery}

Considering the factorization of the precision matrix \eqref{eq:precision}
it is immediate to
consider the following minimization problem for the $\ell_1$-penalized minus
 log-likelihood 
\begin{equation}  \begin{array}{ll} 
	\text{minimize } &  -2\log\det(A) + \operatorname{trace}(A^t\hat{R}A)  
	+  \lambda ||A||_1 \\
	\text{subject to } & A \in \mathbb{R}^{p \times p} \text{ invertible} 
\end{array},
\label{eq:optim}
	\end{equation} 
where $\hat{R}$ is the empirical correlation matrix estimated from the data. 

Solving the above problem, we estimate a sparse factorization of the 
inverse covariance matrix and the estimated graph can be read directly from the 
non-zero entries of the minimizer. 

\begin{remark}
The proposed structural estimation is a general estimation for linear SEM, including
models with cycles. We focus on the recovery of DAGs mainly because the literature 
on learning acyclic graphs is more extense and more methods are available. 
\end{remark}

\subsection{Solving the optimization} 

Various methods can be used to solve  problem~\eqref{eq:optim}, we present here 
a very simple approach using a proximal gradient method~\citep{bach2012,
parikh2014} similar
to the algorithm proposed in \cite{varando2020}. 

The proximal gradient is a method applicable to minimization problems where the 
objective function has a decomposition into a sum of two functions of which one
is differentiable.
In particular, the objective function of problem~\eqref{eq:optim} can be written
as $f(A) + g(A)$ where $f(A) = -2\log\det(A) +
\operatorname{trace}(A^t\hat{R}A)$ and $g(A) = ||A||_1$, with $f$
differentiable. 

The proximal gradient algorithm consists of iterations of the form
\[ A^{(k)} = \mathcal{S}_{s\lambda} \left( A^{(k)} - s\nabla f \right) \]
where the soft-thresholding $\mathcal{S}_{t}(A)_{ij} = \operatorname{sign}(A_{ij}) 
\left(|A_{ij}| - t \right)_+$ is the proximal operator for the
$\ell_1$-penalization and $\nabla f (A) = 2\hat{R} A - 2A^{-1}$. At each iteration the step size $s$ is selected using the 
line search proposed in \cite{beck2010}. 

A complete description of the algorithm  and its implementation is given in the
Appendix~\ref{app:algo}.

\begin{example}
	\label{ex:dag10}

We simulate $1000$ synthetic observations 
	from a Gaussian Bayesian network with associated DAG 
	given in Figure~\ref{fig:example} (a). 
	The graph estimated solving the optimization problem~\eqref{eq:optim} 
	is shown in Figure~\ref{fig:example}~(b).
	We can observe that all the v-structures in the true DAG are correctly 
	recovered. Nevertheless, the estimated graph is not a valid DAG since 
	it contains the directed cycle $7 \leftrightarrow 8$.  
\begin{figure}
	\centering
\begin{tikzpicture}[scale=5]
	\node [node] (v1) at (0.420000, 0.720000)	{1};
	\node [node] (v2) at (0.60000, 0.720000)	{2};
	\node [node] (v3) at (0.80000, 0.720000)	{3};
	\node [node] (v4) at (0.00000, 0.720000)	{4};
	\node [node] (v5) at (0.200000, 0.500000)	{5};
	\node [node] (v6) at (0.500000, 0.340000)	{6};
	\node [node] (v7) at (1.000000, 0.700000)	{7};
	\node [node] (v8) at (1.000000, 0.300000)	{8};
	\node [node] (v9) at (0.00000, 0.340000)	{9};
	\node [node] (v10) at (0.200000, 0.000000)	{10};
	\node (text) at (0.5, -0.3) {(a) True DAG};

	\path [unidirectional] (v1) edge (v5);
	\path [unidirectional] (v4) edge (v5);
	\path [unidirectional] (v5) edge (v9);
	\path [unidirectional] (v5) edge (v10);
 	\path [unidirectional] (v1) edge (v6);
	\path [unidirectional] (v6) edge (v10);
 	\path [unidirectional] (v7) edge (v8);
	\path [unidirectional] (v9) edge (v10);
\end{tikzpicture}
	\hspace{20pt}
\begin{tikzpicture}[scale=5]
	\node [node] (v1) at (0.420000, 0.720000)	{1};
	\node [node] (v2) at (0.60000, 0.720000)	{2};
	\node [node] (v3) at (0.80000, 0.720000)	{3};
	\node [node] (v4) at (0.00000, 0.720000)	{4};
	\node [node] (v5) at (0.200000, 0.500000)	{5};
	\node [node] (v6) at (0.500000, 0.340000)	{6};
	\node [node] (v7) at (1.000000, 0.700000)	{7};
	\node [node] (v8) at (1.000000, 0.300000)	{8};
	\node [node] (v9) at (0.00000, 0.340000)	{9};
	\node [node] (v10) at (0.200000, 0.000000)	{10};
	\node (text) at (0.5, -0.3) {(b) $\lambda = 0.2$};

        \path [unidirectional] (v1) edge (v5);
	\path [unidirectional] (v1) edge (v6);
	\path [unidirectional] (v4) edge (v5);
	\path [unidirectional] (v5) edge (v9);
	\path [unidirectional] (v5) edge (v10);
	\path [unidirectional] (v6) edge (v10);
	\path [bidirectional] (v8) edge (v7);
 	\path [bidirectional] (v7) edge (v8);
	\path [unidirectional] (v9) edge (v10);
\end{tikzpicture}
	\label{fig:example}
	\caption{True DAG (a) and estimated graph with $\lambda=0.2$ (b).}
\end{figure}
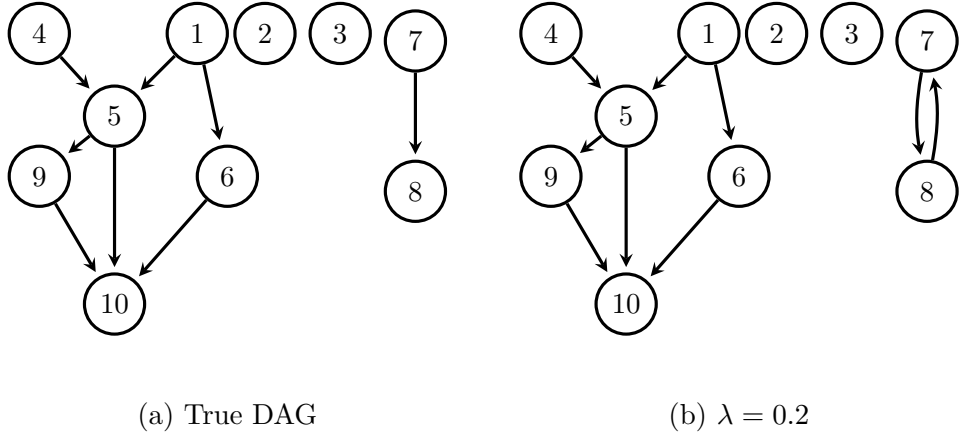
\end{example}

\section{Simulations}
\label{sec:simulations}

We perform a simulation study to explore how the proposed method behaves with respect to the recovery of the 
true graph. 
Data is generated from Gaussian Bayesian networks with known structure
similarly to \cite{colombo14}, in particular random DAGs with $p \in
\{5,10,20,50,100\}$ nodes are generated with independent probability of edges
$\frac{k}{p}$, $k = 1,2,3,4$. Coefficients for the linear regression of each
variable on its parents are independent realizations of a uniform distribution
between $0.1$ and $1$, and the noise distribution is either standard Gaussian or 
exponential with rate parameter equal to $1$. 
For each combination of $p$, $k$ and noise distribution we generate $20$ 
DAGs and subsequently sample $n=100,1000,10000$ observations from the 
induced structural equation models. 

We apply our proposed method (\texttt{nodag}) by solving the optimization
problem~\eqref{eq:optim} with $\lambda = 0.1, 0.2, 0.3$. 
For comparison, we consider three classical structural-recovery 
algorithms: the order independent
PC~\citep{colombo14}, the greedy equivalent search~\citep{chickering2002}, 
and an hill-climbing search.  
The PC algorithm (\texttt{pc}) and the greedy equivalent search (\texttt{ges})
are implemented in the \texttt{pcalg} R package~\citep{pcalg}, while the 
hill-climbing search with tabu (\texttt{tabu}) is available in the 
\texttt{bnlearn} R package~\citep{bnlearn}.
For the \texttt{pc} method we use the Gaussian conditional independence test 
via Fisher's Z and various significance levels ($0.01$, $0.005$ and $0.001$). 
Both \texttt{ges} and \texttt{tabu} methods optimize the Bayesian
information criterion, as default in their implementations. For the 
\texttt{tabu} we also fix the maximum cardinality of the parent set to $10$ 
to limit the computational complexity. 

The pc algorithm and the greedy equivalent search estimate a representation
of the Markov equivalence class while the hill-climbing search and the 
proposed \texttt{nodag} method estimate a directed graph.  

\subsection{Results}

\begin{figure}
	\includegraphics{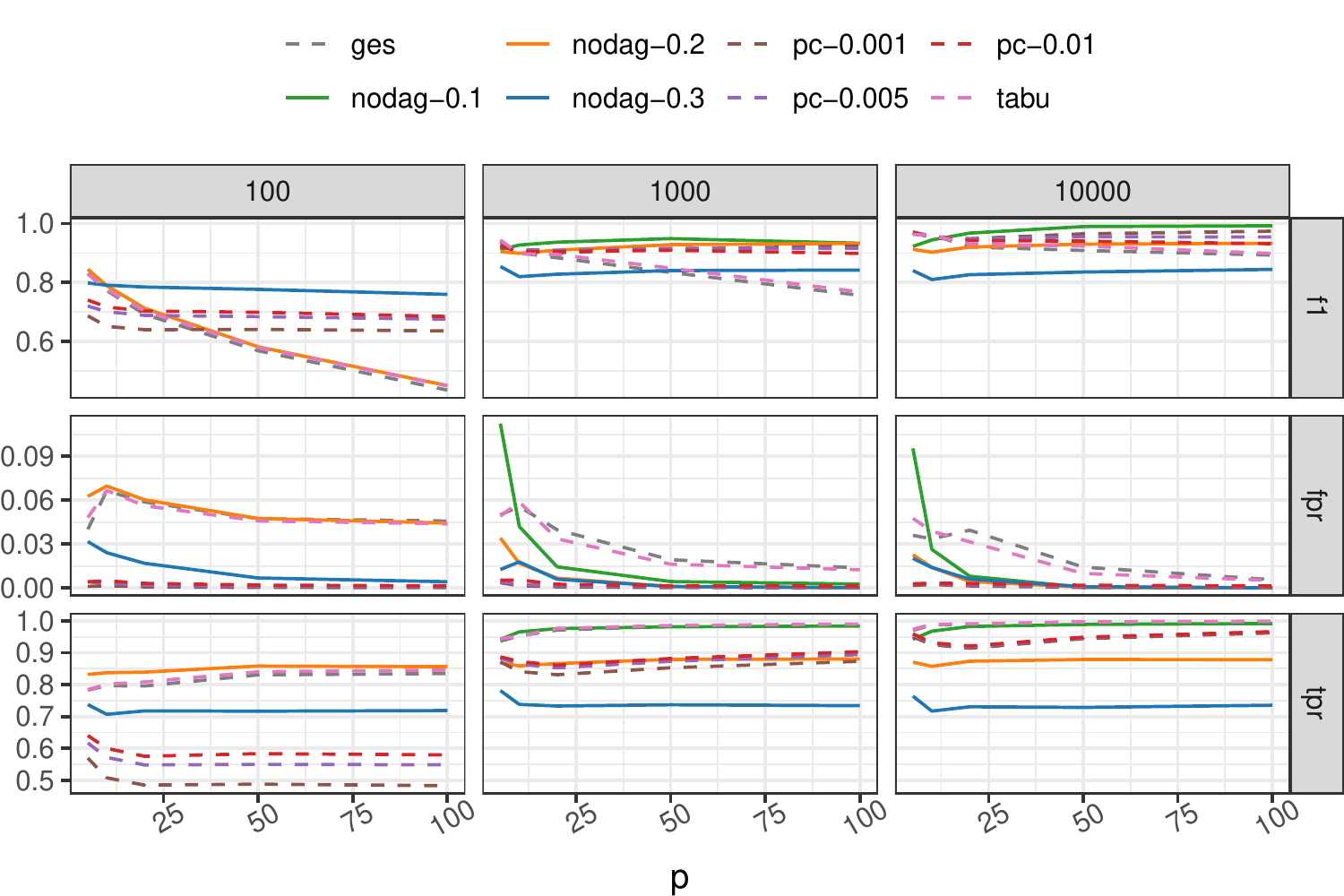}
	\caption{Average F1 score, true positive rate and false positive rate 
	for the recovery of the skeleton.
	Different algorithms in different colours, the proposed \texttt{nodag} method with
	solid lines and algorithms from the literature dashed.
	Results for \texttt{nodag-0.1} are not shown for $n=100$ to improve 
	readability.  
	}
	\label{fig:skeleton}
\end{figure}

Similarly to \cite{colombo14} we evaluate the estimated graphs using the 
F1 score (\texttt{f1}),
false positive
rate (\texttt{fpr}) and true positive rate (\texttt{tpr}) 
with respect to the true skeleton recovery. 
Figure~\ref{fig:skeleton} shows the average metrics for the skeleton recovery 
for the different algorithms. 
We observe that the \texttt{nodag} method obtains, in general, 
comparable results to the literature algorithms, while performing clearly better in the small sample size with respect to skeleton recovery and slightly worse in the large
sample size and small graphs. 

\begin{figure}
	\includegraphics{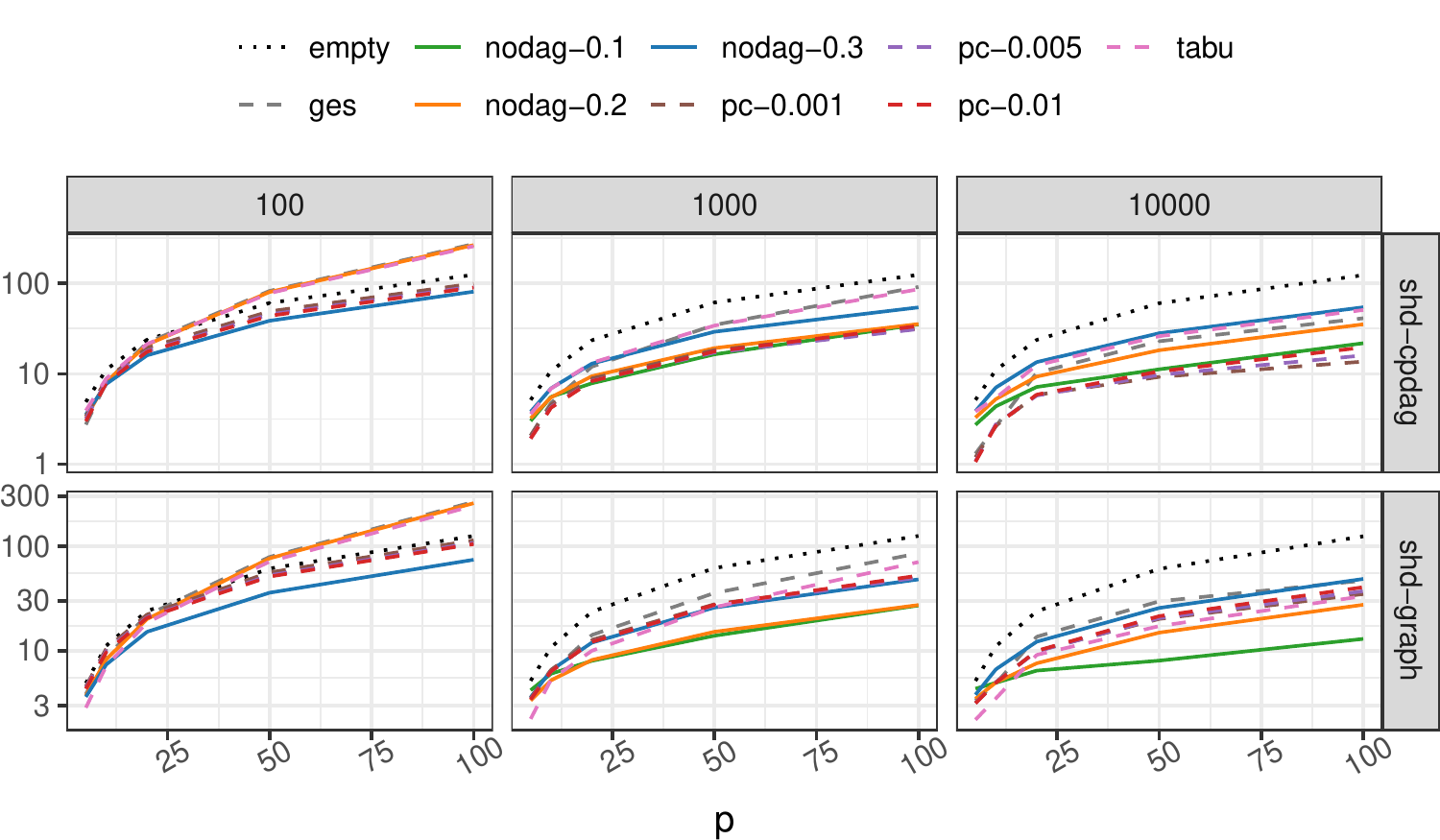}
	\caption{Average structural hamming distances with respect
	to the true CPDAG (\texttt{shd-cpdag}) and to the true 
	DAG (\texttt{shd-graph}).
	Different algorithms in different colours, 
	the proposed \texttt{nodag} method with
	solid lines and algorithms from the literature dashed. The distance with 
	respect to the empty graph is shown with black dotted lines.
	Results for \texttt{nodag-0.1} are not shown for $n=100$ to improve 
	readability.} 
	\label{fig:shd}
\end{figure}

Evaluating edge directions is more tricky, since different DAGs with the same 
skeleton can be Markov equivalent (see Section~\ref{sec:markov}), and moreover
algorithms can output estimated directed or partially directed graphs. 
We chose to report structural Hamming distance to both the true DAG 
(\texttt{shd-graph}) and the true CPDAG (\texttt{shd-cpdag}) in Figure~\ref{fig:shd}. 
We can see that the proposed method is on average superior to other 
algorithms with respect to the recovery of the true DAG.
As for the skeleton recovery, \texttt{nodag} performs worst in the 
large sample and small system dimensions. 
As we showed in Example~\ref{ex:dag10} the graphs estimated by \texttt{nodag}
have,sometimes, a double edge $i\leftrightarrow j$, and we have 
observed that this happens especially when the true CPDAG have the corresponding undirected edge 
$i - j$. 

In general the value of the penalization coefficient $\lambda$ 
obtaining the optimal results is, as expected, dependent on the sample size, but 
on average it does not seem to be too much sensitive. 
  
\begin{figure}[h!]
	\includegraphics{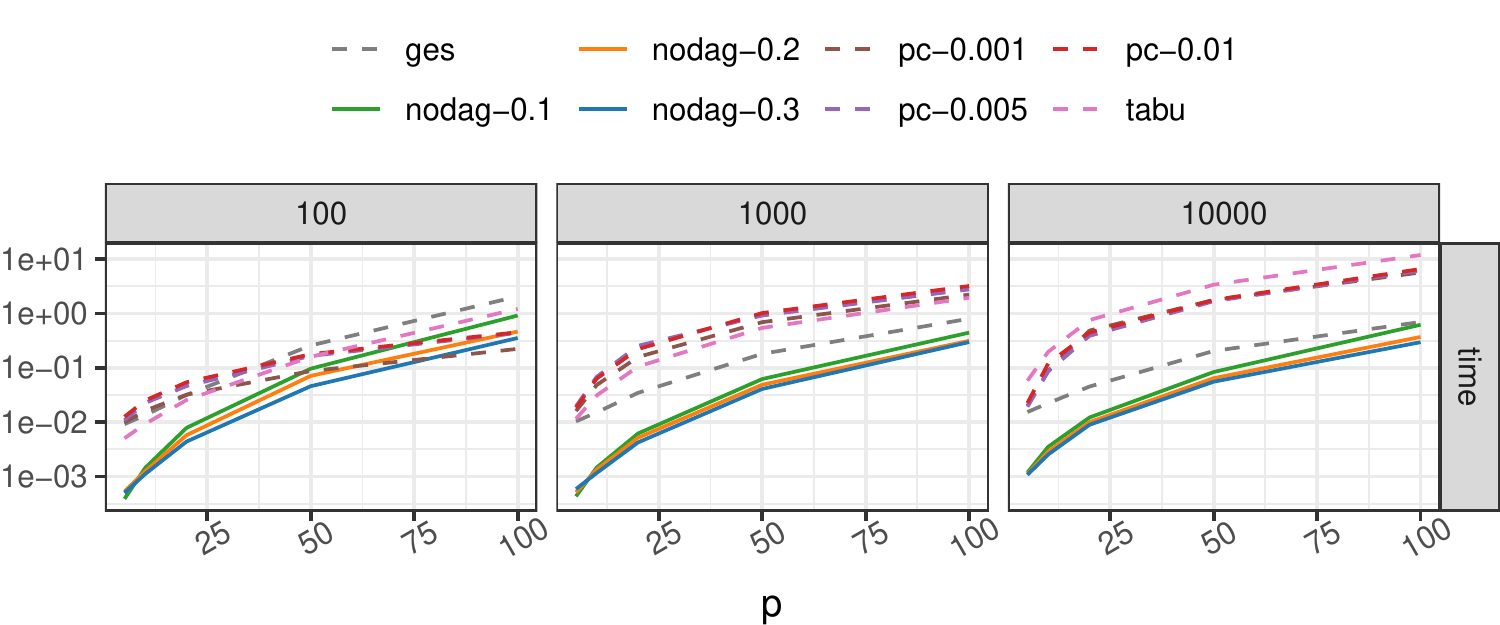}
	\caption{Average execution time
	for different algorithms (colours), the proposed \texttt{nodag} method with
	solid lines and algorithms from the literature dashed.} 
	\label{fig:time}
\end{figure}

In Figure~\ref{fig:time} we report the average execution
time\footnote{Simulations performed on a standard laptop with 8Gb of RAM and 
an i5-8250U CPU} for the different methods.  
The \texttt{nodag} method is shown to outperform all the other methods from a 
computational speed prospective, especially for large sample and system sizes. 

\section{Protein signaling network} 

As an example of real-data application we consider the 
dataset from \cite{sachs2005}  largely
used~\citep{friedman2007, meinshausen2016, zheng2018, varando2020} 
in the graphical models and causal discovery literature. 

Data consist of observations of phosphorylated proteins and 
phospholipids ($p=11$) from cells under different conditions ($n=7466$). 

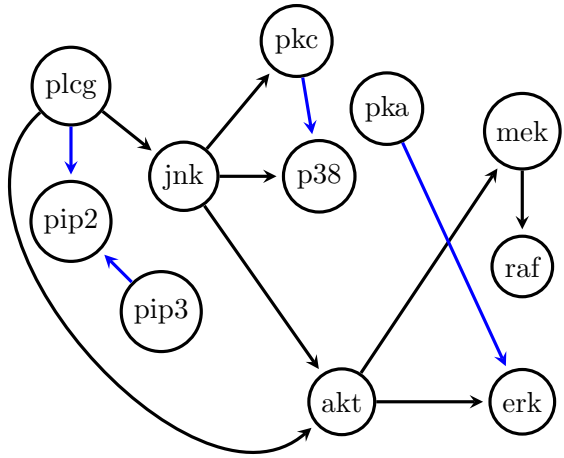
\begin{figure}
\centering
\begin{tikzpicture}[scale=6]
	\node [node] (v1) at (1.000000, 0.300000)	{raf};
	\node [node] (v2) at (1.000000, 0.600000)	{mek};
	\node [node] (v3) at (0.000000, 0.700000)	{plcg};
	\node [node] (v4) at (0.000000, 0.400000)	{pip2};
	\node [node] (v5) at (0.200000, 0.200000)	{pip3};
	\node [node] (v6) at (1.000000, 0.000000)	{erk};
	\node [node] (v7) at (0.600000, 0.000000)	{akt};
	\node [node] (v8) at (0.700000, 0.650000)	{pka};
	\node [node] (v9) at (0.500000, 0.800000)	{pkc};
	\node [node] (v10) at (0.550000, 0.500000)	{p38};
	\node [node] (v11) at (0.250000, 0.500000)	{jnk};

	\path [unidirectional] (v2) edge (v1);
	\path [unidirectional, color = blue] (v3) edge (v4);
	\path [unidirectional, bend right = 90] (v3) edge (v7);
	\path [unidirectional] (v3) edge (v11);
	\path [unidirectional, color = blue] (v5) edge (v4);
	\path [unidirectional] (v7) edge (v2);
	\path [unidirectional] (v7) edge (v6);
	\path [unidirectional, color = blue] (v8) edge (v6);
	\path [unidirectional, color = blue] (v9) edge (v10);
	\path [unidirectional] (v11) edge (v7);
	\path [unidirectional] (v11) edge (v9);
	\path [unidirectional] (v11) edge (v10);
\end{tikzpicture}
	\caption{Graph estimated as the support of $A^*$  from solving the
	optimization problem~\eqref{eq:optim} with $\lambda=0.2$. In blue edges
	that are present in the conventionally accepted network~\citep{sachs2005}.}
	\label{fig:sachs}
\end{figure}

We estimate the graph representing the protein-signaling network 
with our \texttt{nodag} method with $\lambda=0.2$ (Figure~\ref{fig:sachs}). 
The \texttt{nodag} method estimates a graph without cycles and with 
$12$ edges of which $4$ are also present in 
the consensus network~\citep{sachs2005}.
Tow estimated edges, jnk $\rightarrow$ pkc and mek $\rightarrow$ raf, appear with 
reversed direction in the consensus network while others (jnk $\rightarrow$ p38, 
akt $\rightarrow$ erk, akt $\rightarrow$ mek) appear in the graph estimated from other methods in the 
literature~\citep{meinshausen2016, varando2020}. 

\section{Discussion}

We have framed the problem of learning DAGs in the larger class of linear  
structural equation models and we have shown that a simple approach based on optimization techniques and without any acyclicity constraints is able to obtain similar 
recovery performances than state-of-the-art algorithms, see Figures~\ref{fig:skeleton} 
and \ref{fig:shd}.

The proposed \texttt{nodag} is also considerably faster than classical 
constraint-based and score-based methods (Figure~\ref{fig:time}). 
By avoiding the acyclicity constraint we are able to use a standard proximal gradient
algorithm over a matrix $A \in \mathbb{R}^{p \times p}$ and thus the computational
cost of the method depends only on the size of the system ($p$) and not on the 
sparsity level or the sample size. 

Moreover, the output of the \texttt{nodag} method parametrizes the inverse 
covariance matrix and thus provides an estimation of the parameters of the model, 
contrary to constraint-based algorithms. 
In particular, from the estimation of the $A$ matrix is possible to recover an 
estimate of the coefficient matrix $\Lambda$ in equation~\eqref{eq:precision}.
 
\subsection{Feature directions} 

The \texttt{nodag} method estimates linear SEM without acyclicity constraints and thus
it would be interesting to empirically test its performance in the recovery of SEM with cycles. 

As for lasso~\citep{friedman2010}, graphical lasso~\citep{friedman2007} and other $\ell_1$-penalized 
methods~\citep{varando2020} it is straightforward to extend the 
method to estimate the regularization path 
for a sequence of decreasing $\lambda$ values and thus being able to 
perform data-driven selection for the regularization coefficient. 
The good computational complexity make it feasible to combine the proposed algorithm 
with stability selection methods~\cite{meinshausen2010} and, by not imposing 
acyclicity, it allows to simply average matrices estimated from 
bootstrapped or sub-sampled data.

Finally, from a strictly computational point of view, it would be interesting 
to explore ways to speed up both the computations and the convergence of 
Algorithm~\ref{alg:prox}. 
The  sparsity of matrix $A$ in Algorithm~\ref{alg:prox} 
could probably be used to speed up or avoid the computation of its LU 
decomposition, this  would in turn
decrease the cost-per-iteration. 
Accelerated gradient-like methods could be 
applied to improve the speed of convergence~\citep{beck2009}, thus reducing the 
number of iterations needed to converge. 

\subsection{Reproducibility} 

The code and instructions to reproduce the examples, the simulation study and the 
real-world application are available at \url{https://github.com/gherardovarando/nodag_experiments}.

\subsection*{Acknowledgments}
This work was supported by VILLUM FONDEN (grant 13358).

\appendix

\section{Algorithm implementation}

Algorithm~\ref{alg:prox} details the  pseudocode of the proposed proximal 
method to solve problem~\ref{eq:optim}. 
Each iteration of the algorithm consists basically in the gradient computation
and the line search loop where the gradient descent and the proximal operator 
are applied for decreasingly small step sizes until the descent  and 
the \cite{beck2010} conditions are met. The algorithm terminates when the 
difference of the objective function computed in the last two iterations is less
than a specified tolerance ($\varepsilon$) or when the maximum number of iterations
($M$)  has
been reached. 
The LU factorization of $A$, used to compute both the log-determinant and the 
inverse, is performed with the LAPACK implementation~\cite{anderson1999} using
partial pivoting. 

The fortran code implementing algorithm~\ref{alg:prox} is available at 
\href{https://github.com/gherardovarando/nodag}{\texttt{github.com/gherardovarando/nodag}}.

\label{app:algo}
\begin{algorithm}[h!]
	\caption{Proximal gradient algorithm for minimization of
	$\ell_1$-penalized minus log-likelihood}
	\label{alg:prox}
	\begin{algorithmic}[1]
		\small
		\REQUIRE  $\hat{R}$ 
		 the empirical correlation matrix, \\
		\hspace{7pt} $M \in \mathbb{N}$,  
		 $\varepsilon > 0, \lambda > 0 ,  \alpha \in (0,1)$ 
		\STATE initialize $k=0$, $A = L = U =  I$ 
		\STATE{$f =  \sum_{i = 1}^p
		\hat{R}_{ii}$}
		\STATE{$g =  0 $}
	        \REPEAT 
		\STATE increase iteration counter $k = k +1$
		\STATE compute $A^{-1}$ using the LU decomposition 
		\STATE obtain the gradient {${D} = 2\hat{R}A - 2A^{-1}$}
		\STATE copy $A$ into $A'$
		\STATE $f' = f$, $g' = g$
                \STATE{$s = 1$} 
		\LOOP
		\STATE{${A} = {A'} - s  D$ }
		\STATE{soft thresholding $A$ at level $s\lambda$} 
		\STATE $L,U = $ LU decomposition of $A$  
		\STATE compute $\hat{R}A$ 
		\STATE{$f = -2 \sum_{i=1}^p \log(U_{ii}) + \sum_{i,j = 1}^p
		A_{ij} \left(\hat{R} A\right)_{ij}$}		
		\STATE{$g =  \lambda ||A||_1 $}
		\STATE{$\nu = \sum_{i,j}^p \frac{1}{2s}(A_{ij} - A'_{ij})^2 
			  +  (A_{ij} - A_{ij}')D_{ij}$}
                \IF{$f \leq f' + \nu $ and $f + g \leq f' + g'$}
		\STATE{\textbf{break}}
		\ELSE 
		\STATE{$s = \alpha s$}
		\ENDIF
		\ENDLOOP 
		\STATE{$\delta = (f' + g' - f - g) $}
		\UNTIL{$k > M$ \OR $\delta < \varepsilon$}
		\ENSURE   $A$, $f$, $\delta$, $k$ 
	\end{algorithmic}
\end{algorithm}

\vskip 0.2in

\bibliographystyle{plainnat}
\bibliography{biblio}
\end{document}